\documentclass[runningheads]{llncs}

\usepackage{makeidx} 
\usepackage[misc,geometry]{ifsym} 
\usepackage{amsmath} 
\usepackage{bm} 
\usepackage{graphicx} 
\usepackage{array} 
\usepackage[table,xcdraw]{xcolor} 
\usepackage{color} 
\usepackage{afterpage} 
\usepackage[labelfont=bf]{caption} 
\usepackage{footmisc} 
\usepackage[colorlinks]{hyperref} 
\usepackage{fixltx2e} 
\usepackage{cite} 
\usepackage{listings} 
\usepackage{enumitem}
\setlist[itemize]{leftmargin=*} 

\newcommand{\cmmnt}{}

\usepackage{etoolbox}
\makeatletter
\let\llncs@addcontentsline\addcontentsline
\patchcmd{\maketitle}{\addcontentsline}{\llncs@addcontentsline}{}{}
\patchcmd{\maketitle}{\addcontentsline}{\llncs@addcontentsline}{}{}
\patchcmd{\maketitle}{\addcontentsline}{\llncs@addcontentsline}{}{}
\makeatother
\usepackage{hyperref}
\usepackage{bookmark}

\begin{document}

\title{Software-Defined FPGA Accelerator Design for Mobile Deep Learning Applications}

\titlerunning{SW-Defined FPGA Accelerator Design for Mobile DL Applications}  

\author{Panagiotis~G.~Mousouliotis\textsuperscript{(\Letter)}\orcidID{0000-0001-9621-924X},
Loukas~P.~Petrou\orcidID{0000-0001-5760-2043}}

\authorrunning{P. G. Mousouliotis et al.} 

\tocauthor{Panagiotis G. Mousouliotis and Loukas P. Petrou}

\institute{Division of Electronics and Computer Engineering,\\
Department of Electrical and Computer Engineering, Faculty of Engineering,\\
Aristotle University of Thessaloniki, 54124 Thessaloniki, Greece\\
\email{pmousoul@ece.auth.gr}, \email{loukas@eng.auth.gr}}

\maketitle

\begin{abstract}

\cmmnt{Recently, the field of deep learning has received great attention by the scientific community and it is used to provide improved solutions to many computer vision problems.} Convolutional neural networks (CNNs) have been successfully used to attack problems such as object recognition, object detection, semantic segmentation, and scene understanding. The rapid development of deep learning goes hand by hand with the adaptation of GPUs for accelerating its processes, such as network training and inference. Even though FPGA design exists long before the use of GPUs for accelerating computations and despite the fact that high-level synthesis (HLS) tools are getting more attractive, the adaptation of FPGAs for deep learning research and application development is poor due to the requirement of hardware design related expertise. This work presents a workflow for deep learning mobile application acceleration on small low-cost low-power FPGA devices using HLS tools. This workflow eases the design of an improved version of the SqueezeJet accelerator used for the speedup of mobile-friendly low-parameter ImageNet class CNNs, such as the SqueezeNet v1.1 and the ZynqNet. Additionally, the workflow includes the development of an HLS-driven analytical model which is used for performance estimation of the accelerator. \cmmnt{This model can be also used to direct the design process and lead to future design improvements and optimizations.}


\end{abstract}

\section{Introduction}
\label{sec:intro}

HLS tools \cite{nane} provide a higher level of abstraction in digital design and increased productivity when compared to more traditional design methods, such as the hardware description languages (HDLs). This increased productivity comes at the cost of limited design flexibility, compared to HDLs, plus a steep learning curve \cmmnt{\cite{xil_hls, ali}}. \cmmnt{In addition, the successful HLS design of an accelerator does not imply successful deployment and integration in the case where the accelerator is part of a larger FPGA system-on-chip (SoC) design. The development becomes even more complicated if the designer has to call the accelerator inside an operating system (OS), which runs on the CPU part of the FPGA SoC, because this communication requires the development of a low-level kernel driver \cite{gschw}.} To \cmmnt{override these problems and} make HLS-driven design more attractive, Xilinx introduced the SDSoC tool \cite{xil_sdsoc}. With SDSoC, the user marks for FPGA acceleration functions of the input C/C++ application code which are compiled to be run on the CPU side of the FPGA SoC. \cmmnt{Given that the marked for acceleration functions are written in an HLS-compatible way, SDSoC will: (1) use Vivado HLS to produce HDL code for them, (2) use Vivado to implement the accelerator and generate the FPGA bitstream, (3) build the software bare-metal or OS application and the related drivers, and (4) generate sd-card contents which can be used to run the application on a target FPGA SoC board.}

In order to develop mobile deep learning applications on an FPGA SoC, we accelerate mobile-friendly ImageNet class CNNs \cite{iand_small, mous_exp}, which are characterized by small model size and, relatively, limited computational requirements. The characteristics of these CNNs translate in limited requirements in terms of BRAM and DSP FPGA resources; meaning that small, low-power, and low-cost FPGA SoC devices, such as the xc7z020clg484-1 FPGA SoC device, can be used.

Porting mobile-friendly CNNs onto small FPGA SoCs using SDSoC is not a straightforward procedure. Our contribution includes a workflow where: (1) CNNs are first described in a higher-than-C/C++-level language such as Matlab, (2) CNNs' feature-maps and parameters are quantized at 8-bit dynamic fixed-point format using Ristretto \cite{gysel}, (3) the quantized CNNs are implemented in C/C++, (4) the computational intensive functions of the C/C++ description are re-written in a HLS-compatible way in order to be accelerated, and, finally, (5) the SDSoC tool is used for developing an application and deploying it to a specific FPGA SoC board. In this work, we also improve and extent the design of the SqueezeJet \cite{mous_sqj} accelerator and use it to accelerate both SqueezeNet v1.1\footnote{ https://github.com/DeepScale/SqueezeNet/tree/master/SqueezeNet\_v1.1} \cite{iand_sqn} and ZynqNet \cite{gschw} CNNs achieving 13.34 fps for the execution of the SqueezeNet v1.1 and 11.54 fps for the ZynqNet on the xc7z020clg484-1 FPGA SoC device. Finally, we show how the HLS performance estimation information can be used to develop an analytical model of an accelerator design. The results of the analytical model of our accelerator are the closest to the real accelerator latency measurements performed on the FPGA SoC device when compared with the performance estimation and the C/RTL Co-Simulation functionalities of Vivado HLS.

The rest of the paper is organized as follows: Section \ref{sec:sota} presents related work. Our software-defined workflow is described in Section \ref{sec:sdwork}. Section \ref{sec:acc_des} presents the SqueezeJet-2 accelerator design as an improved version of SqueezeJet. The development of our analytical model is presented in Section \ref{sec:hls_anal}. Section \ref{sec:exp_res} shows: (1) results related to the performance of our analytical model in terms of accuracy, and (2) results related to the performance of our accelerator in terms of latency and resources utilization. Finally, Section \ref{sec:con_fu} concludes the paper and proposes future work.

\section{Related work}
\label{sec:sota}

In this section we refer to works that could be used to develop mobile deep learning applications with FPGA SoCs. Mobile computer vision applications (automotive, drones, etc.) often pose real-time performance constraints translating in minimal latency or a batch size equal to one.

ZynqNet describes a CNN architecture and an HLS design for the acceleration of this network. \cmmnt{This approach shows that for achieving the desirable results, the problem must be transformed to aim the processing on the selected platform.} ZynqNet derived from SqueezeNet by replacing the combination of convolutional and maxpool layers with a convolutional layer having increased stride \cite{spring}. This transformation simplifies the accelerator design; by implementing a convolutional layer and a global pooling layer, the ZynqNet accelerator can process the whole CNN except the last softmax layer. Convolutional layer acceleration is achieved by calculating multiple output feature-map channels in parallel using processing elements (PEs) which fully unroll the calculation of a $ 3 \times 3 $ kernel.

In Angel-Eye \cite{guo}, a design flow for mapping CNNs onto embedded FPGA devices is proposed. This design flow includes a dynamic fixed-point quantization strategy, a software controlled hardware architecture with $ 3 \times 3 $ convolution kernel support, and a run-time workflow which allows a single frame to be processed by multiple CNNs. Since real-time processing is of main concern, Angel-Eye uses a batch size of one in order to minimize latency.

In \cite{ven_lat} a latency-driven design method is presented as an extension of the fpgaConvNet modeling framework \cite{ven_fpga}. This work models CNNs using the synchronous dataflow (SDF) model of computation. CNNs are interpreted as directed acyclic graphs (DAGs) whose nodes are mapped to hardware building blocks which are interconnected to form the final SDF graph. The SDF model of computation allows the generation of static schedules of execution and the calculation of the amount of buffer memory between the interconnected hardware building blocks. The SDF graph is partitioned along its depth and a single flexible reference architecture is generated which enables the execution of all the subgraphs. In contrast to Angle-Eye, this reference architecture is tailored to a specific CNN and it is optimized in terms of latency. Their framework produces synthesizable Vivado HLS code for the resulting architecture.

We follow a similar approach to ZynqNet by developing an accelerator which targets CNNs optimized for embedded mobile applications; the architecture of these CNNs can be easily adapted to run on FPGA SoCs. For this purpose, we improve the design of the SqueezeJet convolutional layer accelerator and we also add to it support for performing the maxpool operation. Similarly to Angel-Eye, we use Ristretto for 8-bit dynamic fixed point data quantization. Our accelerator is software controlled and it is using parallel operating PEs that execute concurrently $ 1 \times 1 $ kernel convolutions which calculate multiple output feature map channels. Thus, it can support arbitrary convolution kernel sizes without limiting the utilization of the accelerator computing resources, which are valuable in embedded mobile FPGA SoC devices. We also use a batch size of one to minimize the latency and achieve real-time performance. Finally, we don't use a mathematically-driven design methodology as it is the case with fpgaConvNet, but we derive an analytical model for the performance estimation of our accelerator, which can be used for design improvements by means of design-space exploration.

\section{Software-Defined Workflow}
\label{sec:sdwork}

This workflow could be generalized and applied for the FPGA acceleration of any algorithm if a quantization framework existed which could handle a broad range of algorithms.

We used Ristretto \cite{gysel}, a deep learning quantization framework implemented as a Caffe \cite{jia} extension. Ristretto decreases the bit width of the feature-maps and parameters in every CNN layer and performs a CNN forward pass to get the accuracy; it keeps reducing the bit width up to a network accuracy threshold set by the user. We quantize both the SqueezeNet v1.1 and the ZynqNet CNNs down to 8 bits in both feature-maps and parameters using dynamic fixed-point arithmetic. The top-1 accuracy drop is 2.76\% and 1.44\% for the SqueezeNet v1.1 and the ZynqNet networks respectively.

We adapted and extended a SqueezeNet Matlab project\footnote{ https://github.com/mtmd/SqueezeNet\_MATLAB} to support the forward pass of SqueezeNet v1.1 and ZynqNet in floating-point and dynamic fixed-point modes. Matcaffe, a Caffe Matlab interface, is used to generate the network parameters and inter-layer network data in order to compare the Caffe results against those of the Matlab implementation. The results from Ristretto are used in the Matlab implementation to generate the parameters for a dynamic fixed-point network model; we developed a Matlab script that can be used to save the generated network parameters to binary files.

Furthermore, we developed a C/C++ project which implements and tests the forward pass of the floating-point and fixed-point versions of the CNNs by using the binary files generated in the aforementioned Matlab project.

In the next step, the computationally intensive CNN layer functions, the convolutional and the maxpool, are re-written in an HLS-compatible way.

Finally, the whole C/C++ project is imported in SDSoC for testing and implementation.

\section{Accelerator Design}
\label{sec:acc_des}

\subsection{SqueezeJet-2}

SqueezeJet-2 is an improved re-design and extension of the SqueezeJet accelerator \cite{mous_sqj}; its improvements follow.

\textit{Support for stride values larger than 1:} The ZynqNet CNN uses convolutional layers with a stride equal to 2.

\textit{Single accelerator design:} SqueezeJet used two accelerators; one for the first SqueezeNet v1.1 layer and another one for the rest of the layers. Our current implementation uses a single accelerator for all the CNNs’ layers. To overcome the first layer's small input channel issue, we use a software solution that reshapes the input and the parameters of the first layer in order to increase the computation utilization of our accelerator. \cmmnt{For example, the first layer of SqueezeNet v1.1 is reshaped from ($227\times 227\times3$, 3, 2, 0, 64) to ($113\times113\times32$, 1, 0, 64) zero-padding the new input feature-map where it is required; the meaning of the notation ($113\times113\times32$, 1, 0, 64) is (from left to right) the input feature-map size, kernel size, stride, padding, and number of output channels.}

\textit{Use of double buffering technique:} In this way, the communication latency of reading the input feature-map data is hidden behind the computation of the convolution operation.

\textit{Support for the maxpool operation:} We re-arranged the SqueezeNet v1.1 layers to bring the maxpool layers before the merge layers. The idea is to make it possible for our accelerator to perform the calculations for both the convolutional and the maxpool layers without the need of sending data back to main memory. In the case where only the convolution operation is required, the maxpool operation is bypassed.

\textit{Use of the LUTs to increase the number of the implementable multiply-accumulate (MAC) units:} Although the xc7z020clg484-1 FPGA SoC device includes 220 DSP blocks, we managed to implement 256 MAC units (16 PEs with each of them consisting of 16 MAC units) by making use of the resources related HLS pragma; we implemented half the MAC units using DSP blocks and the other half using LUTs. Unfortunately, the HLS synthesis tool is unable to map multiple 8-bit multiplications on a single DSP block.

\textit{Support for dynamic fixed-point arithmetic.}

\subsection{Cache organization and Operation}

Below, we provide details related to the SqueezeJet-2 cache organization and operation regarding the convolution operation; the description of the caches used for the implementation of the convolution operation follows.

\texttt{\_weights[PAR\_FACT][Q\_CHOxKxKxCHI\_MAX]}: Consists of a group of \texttt{PAR\_FACT} caches of size \texttt{Q\_CHOxKxKxCHI\_MAX}. These caches result from the partitioning of the \texttt{\_weights} array with a factor of \texttt{PAR\_FACT}. This is done in order to have simultaneous access to these caches. Each of these caches is used by one of the \texttt{PAR\_FACT} PEs, which is responsible to calculate \texttt{Q\_CHO} output channels of a specific output pixel\footnote{In this work, ``pixel'' is used to describe the set of all the channels that can be addressed with some specific spatial coordinates. This notion extents in the case of a feature-map line or a line-buffer line.}. Because the weight and bias parameters will be reused in the calculation of every output-feature-map pixel, we store all the CNN parameters in the on-chip BRAMs. In case where this is not possible, we partition the parameters in the output-channel dimension and we calculate specific output-feature-map channels in every accelerator function invocation; in the end we merge the partial results in the output-channel dimension to get the final result. \cmmnt{Even though this method is simple, it has the disadvantage of re-reading the input-feature map pixels in every accelerator function invocation; we use this method because the number of times where this case takes place is limited.}

\texttt{\_bias[PAR\_FACT][Q\_CHO\_MAX]}: This cache group holds the CNN layer's bias values and its use is similar to the one of the \texttt{\_weights} cache group.

\texttt{linebuf[K\_MAX][WIxCHI\_MAX]}: This is a single cache of size \texttt{K\_MAX} by \texttt{WIxCHI\_MAX}. It is described in this way because when the kernel's height by width size is larger than $1 \times 1$, some feature-map lines will be reused as the line-buffer array ``slides down'' the input-feature-map.

\texttt{linebuf\_idx[K\_MAX]}: This is an array used as a ``pointer'' that determines the order of the line-buffer lines as they ``slide down'' the input-feature-map. This is done to avoid having an array of pointers-to-line-buffers because the HLS tool used cannot handle passing to functions pointer-to-pointer-to-arrays arguments. \cmmnt{Also, using a specific pointer-to-pointer in multiple functions, causes the HLS tool to inline all functions that use this specific pointer-to-pointer \cite{gschw}. In our implementation, \texttt{linebuf} and \texttt{linebuf\_idx} are passed as function arguments and the \texttt{linebuf[ linebuf\_idx[ ] ][ ]} notation is used to access the desirable line-buffer line. ``Shifting'' the \texttt{linebuf[][]} down the input-feature-map is done by changing (rotating) the contents of the \texttt{linebuf\_idx} array \cite{mous_sqj}.}

\texttt{linebuf\_win0[KxKxCHI\_MAX]}, \texttt{linebuf\_win1[KxKxCHI\_MAX]}: These are two caches which ``slide horizontally'' on the linebuf cache. We use two line-buffer windows instead of one in order to implement double buffering, which is used for overlapping communication with computation; the process of reading new input-feature-map data from the main memory while executing the convolution operation. \cmmnt{Because the latency of reading input-feature-map values is hidden by the parameter initialization and the output-feature-map calculation, the re-reading of the input-feature-map data in the case where the parameters won't fit the device's BRAMs does not affect the performance of the accelerator; the re-reading introduces additional cost only in terms of power consumption.}

\texttt{out\_pix0[CHO\_MAX]}, \texttt{out\_pix1[CHO\_MAX]}: These are two caches which hold an output-pixel result. We use two of them as part of the double buffering implementation.

Below, listing \ref{lst:conv} shows the top level C/C++ HLS description of the convolution operation; pre-calculation of often-used terms, cache initialization, and HLS pragmas are omitted.

\begin{lstlisting}[label={lst:conv}, caption={Top-level HLS description of the convolution operation}]
// For each output row
L_H_OUT: for ( uint8_t ho = 0; ho != h_out; ho++ ) {
  // Shift line-buffer ``down'' in the input-feature-map.
  // Fill the ``last'' line-buffer line with KxCHI values (K 3D pixels).
  shift_linebuf( fmap_in, &iidx, linebuf, linebuf_idx, WIxCHI, KxCHI, kernel,
    stride, ho, h_out, PADxCHI );
  uint16_t lb_pixel_pt = KxCHI; // last line-buffer pixel ``pointer''
  // Line-buffer window initialization
  init_linebuf_win( linebuf, linebuf_idx, linebuf_win0, KxCHI, KxKxCHI, 0 );
  // line-buffer pixel ``pointers'' for line-buffer windows
  uint16_t pixel_iwp0 = ( SxCHI << 1 );
  uint16_t pixel_iwp1 = SxCHI;
  // For each output pixel (in each output row)
  L_W_OUT: for ( uint8_t wo = 0; wo != w_out; wo++ ) {
    if ( wo%2 == 0 ) {
      // Calc pixel
      pixel_calc( linebuf_win0, _weights, _bias, KxKxCHI,
        Q_CHOxKxKxCHI, Q_CHO, out_pix0, ei, eo, ep );
      // Update line-buffer line and line-buffer window
      update_linebuf_win( fmap_in, &iidx, linebuf, kernel,
        SxCHI, WIxCHI, &lb_pixel_pt, linebuf_idx, linebuf_win1,
        KxCHI, KxKxCHI, &pixel_iwp1, PADxCHI, ho, h_out, stride );
      // Write back to off-chip memory
      write_back( fmap_out, &oidx, ch_out, out_pix1, wo, use_relu );
    }
    else {
      // Calc pixel
      pixel_calc( linebuf_win1, _weights, _bias, KxKxCHI,
        Q_CHOxKxKxCHI, Q_CHO, out_pix1, ei, eo, ep );
      // Update line-buffer line and line-buffer window
      update_linebuf_win( fmap_in, &iidx, linebuf, kernel,
        SxCHI, WIxCHI, &lb_pixel_pt, linebuf_idx, linebuf_win0,
        KxCHI, KxKxCHI, &pixel_iwp0, PADxCHI, ho, h_out, stride );
      // Write back to off-chip memory
      write_back( fmap_out, &oidx, ch_out, out_pix0, wo, use_relu );
    }
  }
  // Write back to off-chip memory leftover pixel
  if ( w_out%2 == 0 )
    write_back( fmap_out, &oidx, ch_out, out_pix1, 1, use_relu );
  else
    write_back( fmap_out, &oidx, ch_out, out_pix0, 1, use_relu );
}
\end{lstlisting}

The \texttt{shift\_linebuf()} function ``shifts'' the line-buffer cache down the input feature map using a \texttt{stride} step. Then, using \texttt{init\_linebuf\_win()}, the first line-buffer window is initialized with line-buffer data. Two line-buffer window ``pointers'' are initialized with line-buffer addresses (indices) that will be used to fill the line-buffer windows with new data. The \texttt{pixel\_calc()}, \texttt{update\_linebuf\_win()}, and the \texttt{write\_back()} functions are executed concurrently taking advantage of the double buffering technique. \cmmnt{The \texttt{pixel\_calc()} function calculates one output-feature-map pixel using one of the two line-buffer windows, the \texttt{update\_linebuf\_win()} function updates with new input-feature-map data the line-buffer cache and updates the data in the one of the two line-buffer windows (the one that is not used in the specific \texttt{pixel\_calc()} function call), and the \texttt{write\_back()} function writes back to the main memory the resulted output-feature-map pixel calculated in the previous \texttt{L\_W\_OUT} loop iteration. In this way, the calculation of a new output-feature-map pixel is overlapped with the data update of one of the line-buffer windows and the write back of the output-feature-map data calculated in the previous  \texttt{L\_W\_OUT} loop iteration.}

In our implementation, the SqueezeJet-2 accelerator exchanges data with the ARM CPU of the xc7z020clg484-1 FPGA SoC device using AXI buses. Specifically the interfaces used are: AXI General Purpose (GP) interface for simple arguments, such as input-feature-map size, kernel size, stride, etc., AXI Accelerator Coherency Port (ACP) for input/output-feature-maps which require to be cache coherent since they are used in CNN layers running in the ARM system, such as the merge layer, and AXI High Performance (HP) Port for weight/bias CNN parameters which don't require cache coherency. In the case of the AXI ACP and HP ports, simple DMAs are used for efficient data movement.

\section{HLS-driven Analytical model}
\label{sec:hls_anal}

We derive our analytical model of performance estimation using HLS information such as pipeline depths, and function/loop call overheads.

The convolution operation's performance can be formulated by describing analytically the cost, in terms of cycle count, of the \texttt{precalc\_terms()} function, the \texttt{init\_caches()} function, and the \texttt{L\_H\_OUT} loop which contains the \texttt{shift\_linebuf()} function, the \texttt{init\_linebuf\_win()} function, and the \texttt{L\_W\_OUT} loop. The \texttt{L\_W\_OUT} loop contains three functions with the dominating one being the \texttt{pixel\_calc()} function.
		
The \texttt{pixel\_calc()} function consists of the \texttt{calc\_ch\_out()} and \texttt{write\_pix()} functions operating in dataflow; dataflow is a function-level pipeline operation mode. Function \texttt{calc\_ch\_out()} calculates one output-future-map pixel by assigning the computation to \texttt{PAR\_FACT} parallel-working PEs, each of them calculating \texttt{Q\_CHOxKxKxCHI = CHOxKxKxCHI/PAR\_FACT} MAC operations, \texttt{CHI\_NUM} operations at each cycle. The analytical description of the performance, in terms of cycle count, of the \texttt{calc\_ch\_out()} function is given by the following equation:
\begin{equation}
\begin{gathered}
\scriptstyle CCO_{CC} = CCO\_DSP\_LUT_{CC} = \\
\scriptstyle (CHO \cdot K \cdot K \cdot CHI) / (PAR_{FACT} \cdot CHI_{NUM}) \\
\scriptstyle + PIPE\_CCO\_DSP\_LUT_{FILL} + CCO\_DSP\_LUT_{OVER}
\end{gathered}
\end{equation}
where $\scriptstyle CHO \cdot K \cdot K \cdot CHI$ is the total number of MACs required to calculate one output pixel (it is also the number of weight parameters; it is equal to output-channels by kernel-height by kernel-width by input-channels), $\scriptstyle PIPE\_CCO\_DSP\_LUT_{FILL}$ is the pipeline fill overhead of the \texttt{calc\_ch\_out\_dsp\_lut()} function's loop; the \texttt{calc\_ch\_out\_dsp\_lut()} function call, which represents the calculation done by a PE, is called by the \texttt{calc\_ch\_out()} function, and $\scriptstyle CCO\_DSP\_LUT_{OVER}$ is the overhead introduced with reading the arguments passed by a calling function. This overhead can be significant in the case where the function is called inside multiple nested loops as it is the case of the convolution operation.
In general, for a pipelined loop the performance equation's form is:
\begin{equation}
\scriptstyle LOOP_{CC} = ( TRIPCOUNT \cdot INITIATION\_INTERVAL ) + ITERATION\_LATENCY
\end{equation}
The accelerator is designed in such a way that forces the $\scriptstyle INITIATION\_INTERVAL $ to be equal to 1 for all the pipelined loops. The $\scriptstyle ITERATION\_LATENCY $ can be translated as the pipeline depth of the specific loop.

Using the above example as a guideline, we calculate the total performance of the SqueezeJet-2 accelerator.

\section{Experiments and Results}
\label{sec:exp_res}

\begin{figure}[]
    \centering
    \fbox{ \includegraphics[scale=0.35]{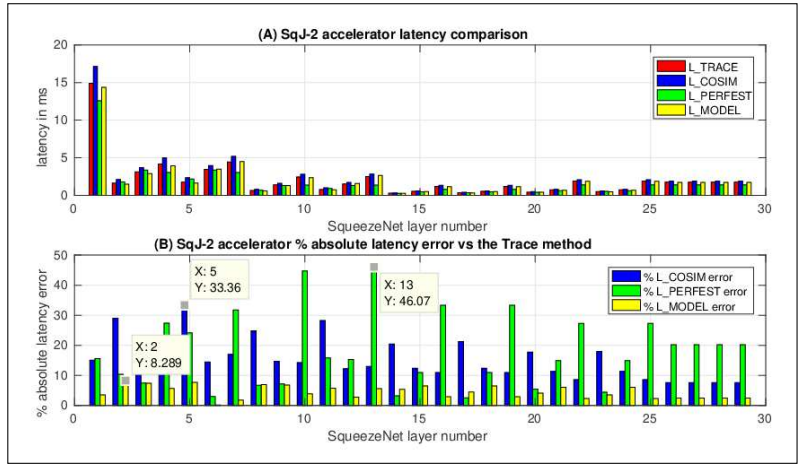} }
    \caption[]{(A) Latency of the accelerator measured with 3 methods and modeled analytically, and (B) the \% absolute latency error of the \texttt{L\_COSIM}, \texttt{L\_PERFEST}, and the \texttt{L\_MODEL} methods, against the \texttt{L\_TRACE} method}
    \label{fig:fig}
\end{figure}

Figure \ref{fig:fig} presents results related to the analytical model of the SqueezeJet-2 dynamic fixed-point (SqJ-2-dfp) accelerator. \texttt{L\_TRACE} represents the latency measurements of the accelerator layers using the hardware tracing feature of SDSoC when the network runs on the FPGA, \texttt{L\_COSIM} represents the latency measurements of the accelerator layers using the C/RTL Co-Simulation feature of the SDSoC, \texttt{L\_PERFEST} represents the worst case latency estimation of the accelerator layers using Vivado HLS synthesis; we explicitly set the min/max tripcounts for the loops of every layer, and \texttt{L\_MODEL} represents the latency estimation of the accelerator layers using the analytical model. From figure \ref{fig:fig} we conclude the following: (1) our analytical model is the closest to the \texttt{L\_TRACE} results, (2) the \texttt{L\_PERFEST} method is the most optimistic and shows up to 46.1 \% error against the \texttt{L\_TRACE} results; the \texttt{L\_COSIM} presents the next max error which is 33.4 \%, and finally our analytical model has a max error of 8.3 \%, and (3) the average \% error of the analytical model of the SqJ-2-dfp accelerator is bellow 5\% (~4.45\%).

Table \ref{table} (A) shows the resources usage of the SqJ-2-dfp, the SqueezeJet-2 floating point (SqJ-2-flp), and the ZynqNet \cite{gschw} floating-point (ZqN-flp) design implementations. Table \ref{table} (B) shows the SqJ-2-dfp, the SqJ-2-flp, and the ZqN-flp accelerators' performance in terms of latency using SqueezeNet v1.1 and ZynqNet as test cases. Table \ref{table} (A) shows that, with the exception of the BRAMs, the SqJ-2-flp accelerator uses almost half the resources used by the ZqN-flp accelerator and table \ref{table} (B) shows that the SqJ-2-flp accelerator is ten times faster than ZqN-flp when executing the ZynqNet CNN. Finally, table \ref{table} (B) shows that the SqN-2-dfp accelerator achieves 13.34 fps for the execution of the SqueezeNet v1.1 and 11.54 fps for ZynqNet on the xc7z020clg484-1 FPGA SoC device.

\begin{table}[]
\centering
\caption{(A) Resources usage of the SqJ-2-dfp, SqJ-2-flp, and the ZqN-flp accelerators; the numbers in parentheses show the \% device resource utilization, and (B) total CNN latency (ms) for the SqJ-2-dfp, SqJ-2-flp, and the ZqN-flp accelerators running at 100MHz; the ZynqNet ZqN-flp result (*) produced using HLS C/RTL Co-Simulation}
\label{table}
\scalebox{0.7}{
\begin{tabular}{|l|r|r|r|}
\hline
\multicolumn{4}{|c|}{\cellcolor[HTML]{AAAAAA}\textbf{(A) Resources Usage}}                                                                                                                                                                                                                                                         \\ \hline
                         & \multicolumn{1}{c|}{\textbf{\begin{tabular}[c]{@{}c@{}}SqJ-2-dfp\\ xc7z020\end{tabular}}} & \multicolumn{1}{c|}{\textbf{\begin{tabular}[c]{@{}c@{}}SqJ-2-flp\\ xc7z045\end{tabular}}} & \multicolumn{1}{c|}{\textbf{\begin{tabular}[c]{@{}c@{}}ZqN-flp\\ xc7z045\end{tabular}}} \\ \hline
\textbf{LUT}             & 36.2k \textit{(68\%)}                                                                              & 63k \textit{(29\%)}                                                                                & 154k \textit{(70\%)}                                                                             \\ \hline
\textbf{LUTRAM}          & 3.1k \textit{(18\%)}                                                                               & 8.8k \textit{(13\%)}                                                                               & ?                                                                                       \\ \hline
\textbf{FF}              & 24.9 \textit{(24\%)}                                                                               & 75.6k \textit{(17\%)}                                                                              & 137k \textit{(31\%)}                                                                             \\ \hline
\textbf{BRAM}            & 96.5 \textit{(69\%)}                                                                               & 324.5 \textit{(60\%)}                                                                              & 498 \textit{(91\%)}                                                                              \\ \hline
\textbf{DSP}             & 172 \textit{(78\%)}                                                                                & 268 \textit{(30\%)}                                                                                & 739 \textit{(82\%)}                                                                              \\ \hline
\multicolumn{4}{|c|}{\cellcolor[HTML]{AAAAAA}\textbf{(B) Total CNN latency (ms)}}                                                                                                                                                                                                                                                  \\ \hline
\textbf{SqueezeNet v1.1} & 74.91                                                                                     & -                                                                                         & -                                                                                       \\ \hline
\textbf{ZynqNet}         & 86.62                                                                                     & *186.8                                                                                    & 1955                                                                                    \\ \hline
\end{tabular}
}
\end{table}

\section{Conclusion and Future Work}
\label{sec:con_fu}

In this work we have demonstrated a workflow which eases the mapping of mobile-friendly CNNs onto low-cost low-power small FPGA SoC devices. We presented an improved version of the SqueezeJet accelerator which achieves 13.34 fps for the execution of the SqueezeNet v1.1 and 11.54 fps for the ZynqNet on the xc7z020clg484-1 FPGA SoC device. Using HLS performance estimation information, we formed an analytical performance estimation model which provides improved performance estimation when compared with the HLS build-in performance estimation and C/RTL Co-Simulation functionalities. Finally, we used C/RTL Co-Simulation and a floating-point version of our accelerator to estimate its performance for the execution of the floating-point version of ZynqNet. The results show that our accelerator is 10 times faster and, with the exception of the BRAMs, uses almost half the FPGA resources when compared against the ZqN-flp accelerator. Future work could use our analytical model for performing design space exploration and optimizing the design of our accelerator.

%
%
\bibliographystyle{splncs04}
\bibliography{mybibfile}

\end{document}